\documentclass[runningheads]{llncs}
\usepackage{array}
\usepackage{adjustbox}
\usepackage{cite}
\usepackage{soul}
\usepackage{url}
\usepackage[utf8]{inputenc}
\usepackage[small]{caption}
\usepackage{graphicx}
\usepackage{amsmath}
\usepackage{latexsym,amssymb}
\usepackage{booktabs}
\usepackage{subcaption}
\usepackage{algorithm2e}
\usepackage[dvipsnames]{xcolor}
\usepackage[english]{babel}
\usepackage{xspace}
\usepackage{xfrac}
\usepackage{setspace}
\usepackage{stmaryrd}
\usepackage{enumitem}
\usepackage{verbatim}
\usepackage[all]{xy}
\usepackage{multirow}

\let\subparagraph\paragraph
\usepackage{titlesec}
\let\subparagraph\llncssubparagraph

\definecolor{tyellow1}{HTML}{FCE94F}
\definecolor{tyellow2}{HTML}{EDD400}
\definecolor{tyellow3}{HTML}{C4A000}
\definecolor{torange1}{HTML}{FCAF3E}
\definecolor{torange2}{HTML}{F57900}
\definecolor{torange3}{HTML}{C35C00}
\definecolor{tbrown1}{HTML}{E9B96E}
\definecolor{tbrown2}{HTML}{C17D11}
\definecolor{tbrown3}{HTML}{8F5902}
\definecolor{tgreen1}{HTML}{8AE234}
\definecolor{tgreen2}{HTML}{73D216}
\definecolor{tgreen3}{HTML}{4E9A06}
\definecolor{tblue1}{HTML}{729FCF}
\definecolor{tblue2}{HTML}{3465A4}
\definecolor{tblue3}{HTML}{204A87}
\definecolor{tpurple1}{HTML}{AD7FA8}
\definecolor{tpurple2}{HTML}{75507B}
\definecolor{tpurple3}{HTML}{5C3566}
\definecolor{tred1}{HTML}{EF2929}
\definecolor{tred2}{HTML}{CC0000}
\definecolor{tred3}{HTML}{A40000}
\definecolor{tlgray1}{HTML}{EEEEEC}
\definecolor{tlgray2}{HTML}{D3D7CF}
\definecolor{tlgray3}{HTML}{BABDB6}
\definecolor{tdgray1}{HTML}{888A85}
\definecolor{tdgray2}{HTML}{555753}
\definecolor{tdgray3}{HTML}{2E3436}

\usepackage[pdftex%
,colorlinks=true       
,linkcolor=tblue2        
,citecolor=tblue2        
,filecolor=black       
,urlcolor=tblue2         
,linktoc=page           
,plainpages=false]{hyperref}
\addto\extrasenglish{%

}

\graphicspath{{plots/}}
\DeclareGraphicsExtensions{.pdf}

\fontfamily{ptm}

\title{Computing Optimal Decision Sets with SAT}

\author{Jinqiang Yu \and Alexey Ignatiev \and Peter J. Stuckey \and Pierre {Le Bodic}}
\authorrunning{Yu et al.}
\institute{
   Faculty of Information Technology, Monash University, Australia \\
   \href{mailto:jyuu0044@student.monash.edu}{\texttt{jyuu0044@student.monash.edu}} \\
   \{\mailtodomain{alexey.ignatiev}\texttt{,}\mailtodomain{peter.stuckey}\texttt{,}\mailtodomain{pierre.lebodic}\}\texttt{@monash.edu}
}
\date{April 2020}

\newcommand{\pjs}[1]{\textcolor{blue}{\textsc{Pjs:} #1}}

\newcommand{\plb}[1]{\textcolor{blue}{\textsc{plb:} #1}}
\newcommand{\ignore}[1]{}
\renewcommand{\plb}[1]{}

\newcommand{\fml}[1]{{\mathcal{#1}}}

\newcommand{\mailtodomain}[1]{\href{mailto:#1@monash.edu}{\texttt{\nolinkurl{#1}}}}

\makeatletter
\def\thm@space@setup{%
  \thm@preskip=2pt \thm@postskip=2pt
}
\makeatother

\titleformat{\paragraph}[runin]
            {\normalfont\normalsize\bfseries}{\theparagraph}{0.4em}{}

\graphicspath{{plots/}}
\DeclareGraphicsExtensions{.pdf}

\begin{document}

\maketitle

\begin{abstract}
  As machine learning is increasingly used to help make decisions,
  there is a demand for these decisions to be \emph{explainable}.
  Arguably, the most explainable machine learning models use decision rules.
  This paper focuses on decision sets, a type of model with unordered
  rules, which explains each prediction with a single rule.
  In order to be easy for humans to understand, these rules must be
  concise.
  Earlier work on generating optimal decision sets first minimizes the
  number of rules, and then minimizes the number of literals,
  but the resulting rules can often be very large.
  Here we consider a better measure, namely the total size of the decision
  set in terms of literals. So we are not driven to a small set of rules
  which require a large number of literals.
  We provide the first approach to determine minimum-size decision
  sets that achieve minimum empirical risk
  and then investigate sparse alternatives where we trade
  accuracy for size.
  By finding optimal solutions we show we can build decision set
  classifiers that are almost as accurate as the best
  heuristic methods, but far more
  concise, and hence more explainable.
\end{abstract}

\section{Introduction} \label{sec:intro}

The world has been changed by recent rapid advances in machine
learning.
Decision tasks that seemed well beyond the capabilities of artificial
intelligence have now become commonly solved using machine
learning~\cite{jordan-science15,bengio-nature15,silver-nature15}.
But this has come at some cost.
Most machine learning algorithms are opaque, unable to explain why
decisions were made.
Worse, they can be biased by their training data, and behave poorly
when exposed to data outside that which they were trained on.
Hence the rising interest in \emph{explainable artificial
intelligence}
(XAI)~\cite{muller-jmlr10,guestrin-kdd16,kim-corr17,lundberg-nips17,guestrin-aaai18,darwiche-ijcai18,rudin-aaai18,muller-dsp18,lipton-cacm18,monroe-cacm18,grefenstette-jair18,inms-aaai19,inms-nips19,guidotti-acmcs19,guidotti-ieee-is19,darwiche-pods20,ignatiev-ijcai20},
including research programs~\cite{gunning-aimag19,ai-roadmap} and
legislation~\cite{eu-reg16,goodman-aimag17}.

In this paper we will focus on classification problems, where the input is a set of \emph{instances} with \emph{features} and, as a label, a \emph{class} to predict.
For these problems, some of the most explainable forms of machine learning formalism are \emph{decisions sets}~\cite{clark-ml89,clark-ewsl91,leskovec-kdd16,ipnms-ijcar18,meel-cp18,dash-nips18,meel-aies19}.
A decision set is a set of decision \emph{rules}, each with conditions $C$ and decision $X$, such that if an instance satisfies $C$, then its class is predicted to be $X$.
An advantage of decision sets over the more popular decision trees and decision lists is that each rule may be understood independently, making this formalism one of the easiest to explain.
Indeed, in order to explain a particular decision on instance $D$,
we can just refer to a single decision rule $C \Rightarrow X$ s.t.\ $D$
satisfies $C$.

For decision sets to be clear and explainable to a human, individual rules should be concise.
Previous work has examined building decision sets which involve the fewest
possible rules, and then minimizes the number of literals in the
rules~\cite{ipnms-ijcar18};
or building a CNF classifer that fixes the number of rules,
and then minimizes the number of literals~\cite{meel-cp18,meel-aies19}
to explain the positive instances of the class.
This work also suffers from the limitation that the rules only
predict class 1, and the model predicts class 0 if no rule applies.
Unfortunately, in order to explain a class 0 instance, we need to use
(the negation of) all rules, making the explanations not succinct.

In this work we argue the number of rules
is the wrong measure of explainability,
since, for example, 3 rules each involving 100 conditions
are most likely less comprehensible than, say,
5 rules each involving 20 conditions.
Indeed, since the explanation of a single instance is just
a single decision rule, the number of rules is nowhere near as
important as the size of the individual rules.
So previous work on building minimum-size decision sets
has not used the best measure of size for explainability.

In this work we examine directly constructing decision sets of the smallest total
size, where the size of a rule with conditions $C$ is
$|C| + 1$ (the additional 1 is for the class descriptor $X$).
This leads to smaller decision sets (in terms of literals) which seem far
more appealing for explaining decisions.

It turns out that this definition of size leads to SAT models that are experimentally harder to solve,
but the resulting decision sets can be significantly smaller.
However, for \emph{sparse decision sets},
where we are allowed to consider a smaller rule set if it does not make too many errors in the classification, this new measure is no more difficult to compute than the traditional rule count measure, and gives finer granularity decisions on sparseness.

The contributions of this paper are
\begin{itemize}
    \item The first approach to building optimal decision sets in terms of
      the total number of literals required to define the entire set,
    \item Alternate SAT and MaxSAT models to tackle this problem, and sparse variations which allow an accuracy versus size trade-off,
    \item Detailed experimental results showing the applicability of this
      approach, which demonstrate that our best approach can generate optimal
      sparse decision sets quickly with accuracy comparable to the best
      heuristic methods, but much smaller.
\end{itemize}

The paper is organized as follows.
\autoref{sec:prelim} introduces the notation and definitions used
throughout the paper.
Related work is outlined in \autoref{sec:relw}.
\autoref{sec:enc} describes the novel SAT- and MaxSAT-based encodings
for the inference of decision sets.
Experimental results are analyzed in \autoref{sec:res}.
Finally, \autoref{sec:conc} concludes the paper.

\section{Preliminaries} \label{sec:prelim}

\paragraph{Satisfiability and Maximum Satisfiability.}
We assume standard definitions for propositional satisfiability
(SAT) and maximum satisfiability (MaxSAT) solving~\cite{sat-handbook09}.
A propositional formula is said to be in \emph{conjunctive}
\emph{normal form} (CNF) if it
is a conjunction of clauses. 
A \emph{clause} is a disjunction of literals. 
A \emph{literal} is either a Boolean variable or its negation.
Whenever convenient, 
clauses are treated as sets of
literals.
Moreover, the term \emph{clausal} will be used to denote formulas
represented as sets of sets of literals, i.e.\ in CNF. 
A truth assignment maps each variable to $\{0,1\}$.
Given a truth assignment, a clause is satisfied if at least one of its
literals is assigned value 1; otherwise, it is falsified.
A formula is satisfied if all of its clauses are satisfied; otherwise,
it is falsified.
If there exists no assignment that satisfies a CNF formula $\fml{F}$,
then $\fml{F}$ is \emph{unsatisfiable}.

In the context of unsatisfiable formulas, the maximum satisfiability
(MaxSAT) problem is to find a truth assignment that maximizes the
number of satisfied clauses.
A number of variants of MaxSAT
exist~\cite[Chapter~19]{sat-handbook09}.
Hereinafter, we will be mostly interested in Partial Weighted MaxSAT,
which can be formulated as follows.
The formula can be represented as a conjunction of \emph{hard} clauses
(which must be satisfied) and \emph{soft} clauses (which represent a
preference to satisfy those clauses) each with a weight.
Whenever convenient, a soft clause $c$ with weight $w$ will be denoted
by $(c, w)$.
The Partial MaxSAT problem consists in finding an assignment that
satisfies all the hard clauses and maximizes the total weight of
satisfied soft clauses.


\paragraph{Classification Problems and Decision Sets.}
We follow the notation used in earlier
work~\cite{bessiere-cp09,leskovec-kdd16,nipms-ijcai18,ipnms-ijcar18}.
Consider a set of features $\fml{F}=\{f_1,\ldots,f_K\}$.
All the features are assumed to be binary (non-binary and numeric
features can be mapped to binary features using standard
techniques~\cite{scikitlearn}).
Hence, a \emph{literal} on a feature $f_r$ can be represented as $f_r$
(or $\neg{f_r}$, resp.), denoting that feature $f_r$ takes value 1
(value 0, resp.).
The complete space of feature values (or \emph{feature
space}~\cite{han-bk12}) is $\fml{U}\triangleq\prod_{r=1}^{K}\{f_r,\neg
f_r\}$.

A standard classification scenario is assumed, in which one is given
training data $\fml{E}=\{e_1,\ldots,e_M\}$.
Each data instance (or \emph{example}) $e_i\in\fml{E}$ is a 2-tuple
$(\pi_i,c_i)$ where $\pi_i\in\fml{U}$ is a set of feature values and
$c\in\fml{C}$ is a class.
(This work focuses on binary classification problems, i.e.
$\fml{C}=\{0,1\}$ but the proposed ideas are easily extendable to the
case of multiple classes.)
\ignore{ A literal $l_r$ on a feature $f_r$,
    $l_r\in\{f_r,\neg{f_r}\}$, is said to \emph{discriminate} an
    example $e_i$ iff feature $f_r$ in example $e_i$ takes the value
    opposite to $l_r$, i.e. $\pi_i[r]=\neg{l_r}$.  \plb{This is not
clear to me. We don't use the term "discriminate" in the paper
anyway.} } An example $e_i$ can be seen as \emph{associating} a set of
feature values $\pi_i$ with a class $c_i\in\fml{C}$.
Moreover, we assume without loss of generality in our context that
dataset $\fml{E}$ \emph{partially} defines a Boolean function
$\phi:\fml{U}\rightarrow\fml{C}$, i.e.\ there are no two examples
$e_i$ and $e_j$ in $\fml{E}$ associating the same set of feature
values with the opposite classes.
(Any two such examples can be removed from a dataset, incurring an
error of 1.)

\ignore{
-- in other words, $(\forall
r\in[K]\ \pi_i[r]=\pi_j[r])\Rightarrow c_i=c_j$ (alternatively and in
order to deal with \emph{inconsistencies}, one can assume $\phi$ to
be a relation defined on subset of $\fml{U}\times\fml{C}$).
}

The objective of classification in machine learning is to devise a
function $\hat{\phi}$ that matches the actual function $\phi$ on the
training data $\fml{E}$ and generalizes \emph{suitably well} on unseen
test data~\cite{furnkranz-bk12,han-bk12,mitchell-bk97,quinlan-bk93}.
%
In many settings (including sparse decision sets), function
$\hat{\phi}$ is not required to match $\phi$ on the complete set of
examples $\fml{E}$ and instead an \emph{accuracy} measure is
considered;
this imposes a requirement that $\hat{\phi}$ should be a relation
defined on $\fml{U}\times\fml{C}$.
Furthermore, in classification problems one conventionally has to deal
with an optimization problem, to optimize either with respect to the
complexity of $\hat{\phi}$,
or with respect to the accuracy of the learnt function (to
make it match the actual function $\phi$ on a maximum number of
examples), or both.

\ignore{
    work has considered various representations of the target function
    $\hat{\phi}$.
Some of the representations are known to be $\ldots$ \cite{citations}.
\pjs{FIX THIS!!}
}
This paper focuses on learning representations of $\hat{\phi}$
corresponding to \emph{decision sets} (DS).
A decision set is an \emph{unordered set} of \emph{rules}.
For each example $e\in\fml{E}$, a rule of the form $\pi \Rightarrow c,
c \in \fml{C}$ is interpreted as \emph{if the feature values of $e$
agree with $\pi$ then the rule predicts that $e$ has class $c$}.
Note that as the rules in decision sets are unordered, it is often the
case that some rules may \emph{overlap}, i.e.\ multiple rules may
agree with an example $e\in\fml{E}$.

\begin{example}\label{ex:data}
Consider the following set of 8 items (shown as columns)

\newcommand{\blue}{\color{tblue3}}
\newcommand{\purple}{\color{tred3}}
\newcommand{\green}{\color{tgreen3}}

\setlength{\tabcolsep}{3pt}
\begin{center}
\begin{tabular}{crrrrrrrrrr}
\toprule
\multicolumn{2}{c}{Item No.} & & 1 & 2 & 3 & 4 & 5 & 6 & 7 & 8 \\
\midrule
\multirow{4}{0mm}{\rotatebox{90}{Features}}
& $L$      & & \blue 1 & \blue 1 & \green 0 & \blue 1 & \green 0 & \blue 1 & 0
& \green 0 \\
& $C$      & & 0 & 0 & \green 0 & \purple 1 & \green 0 & \purple 1 & \purple 1 & \green 0 \\
& $E$      & & 1 & 0 & 1 & 0 & 0 & 1 & 1 & 1 \\
& $S$      & & 0 & 1 & 0 & 0 & 1 & 1 & 0 & 1 \\ \midrule 
\multicolumn{2}{r}{Class $H$}
           & & \blue 0 & \blue 0 & \green 1 & \blue 0 & \green 1 & \blue 0 &
\purple 0
& \green 1 \\
\bottomrule
\end{tabular}
\end{center}

\ignore{
\centerline{
\begin{tabular}{r|rrrr|r}
         & \multicolumn{4}{c}{Features} & Class \\
Item No. &  $L$ & $C$ & $E$ & $S$ & $H$ \\ \hline
    1 &  1 & 0& 1& 0& 0 \\
    2 &  1& 0& 0& 1& 0 \\
    3 &  0& 0& 1& 0& 1 \\
    4 &  1& 1& 0& 0& 0 \\
    5 &  0& 0& 0& 1& 1 \\
    6 &  1& 1& 1& 1& 0 \\
    7 &  0& 1& 1& 0& 0 \\
    8 &  0& 0& 1& 1& 1
\end{tabular}
}
}
A valid decision set for this data for the class $H$ is
\begin{eqnarray*}
\blue L & \blue \Rightarrow & \blue \neg H \\
\green \neg L \wedge \neg C & \green \Rightarrow & \green H \\
\purple C & \purple \Rightarrow & \purple \neg H
\end{eqnarray*}
The \emph{size} of this decision set is 7 (one for each literal on the left
hand and right hand side, or alternatively, one for each literal on the left
hand side and one for each rule). Note how rules can overlap, both the first
and third rule classify items 4 and 6.
\qed
\end{example}

\ignore{
A few additional assumption are made in the following text.
First, we interchangeably use the term \emph{node} instead of
\emph{literal} if necessary.
\plb{The previous sentence is unclear.}
Also, hereinafter and whenever convenient, we treat the \emph{class}
attribute of an example as a distinct feature, i.e.\ we consider
example $e_i\in\fml{E}$ to be a set of feature values $\{f_r\ |\
f_r\in\fml{F}\cup\fml{C}\}$.
Furthermore, such class feature can in the following be denoted by
$c\in\fml{F}\cup\fml{C}$.
}

\section{Related Work} \label{sec:relw}
Interpretable decision sets are a rule-based predictive model that can
be traced at least to~\cite{clark-ml89,clark-ewsl91}.
To the best of our knowledge, the first logic-based approach to the
problem of decision set inference was proposed in~\cite{resende-mp92}.
Concretely, this work proposed a SAT model for synthesizing a formula
in disjunctive normal form that matches a given set of training
samples, which is then tackled by the interior point approach.
Later, \cite{leskovec-kdd16} considered decision sets as a more
explainable alternative to decision trees \cite{breiman-bk84} and
decision lists \cite{rivest-ml87}.
The method of~\cite{leskovec-kdd16} yields a set of rules and
heuristically minimizes a linear combination of criteria such as the
number of rules, the maximum size of a rule, the overlap of the rules,
and error.

The closest related work that produces decision sets as defined in
\cite{clark-ewsl91} is by Ignatiev~\textit{et
al.}~\cite{ipnms-ijcar18}.
Here the authors construct an iterative SAT model to learn minimal, in
terms of number of rules, perfect decision sets, that is where the
decision set agrees perfectly with the training data (which is assumed
to be consistent).
Afterwards, they lexicographically minimize the total number of
literals used in the decision set.
As will be shown later, they generate larger decision sets than our
model, which minimizes the total size of the target decision set.
Their approach is more scalable for solving the perfect decision set
problem, since the optimization measure in use, i.e.\ the number of
rules, is more coarse-grained.
The SAT-based approach of~\cite{ipnms-ijcar18} was also shown to
extensively outperform the heuristic approach
of~\cite{leskovec-kdd16}.

In~\cite{meel-cp18}, the authors define a MaxSAT model for binary
classification, where the number of rules is fixed, and the size of
the model is measured as the total number of literals across all
clauses.
Rather than build a perfect binary classifier, they consider a model
that minimizes a linear combination of size and Hamming loss, to
control the trade-off between accuracy and intrepretability.
The scalability of this approach is improved in \cite{meel-aies19},
where rules are learned iteratively on partitions of the training set.
Note that they do not create a \emph{decision set} as defined in
\cite{clark-ewsl91,leskovec-kdd16,ipnms-ijcar18}, but rather a single
formula that defines the positive instances.
The negative instances are specified by default as the instances not
captured by the positive formula.
This limits their approach to binary classification, and also makes
the representation smaller.
%
For example, on the data of \autoref{ex:data} (and assuming the target
number of rules was 1) they would produce the decision set comprising
a single rule $\neg L \wedge \neg C$.
It also means the explainability is reduced for negative instances,
since we need to use the (negation of the) entire formula to explain
their classification.

Integer Programming (IP) has also been used to create optimal
rule-based models which only have positive rules.
In \cite{dash-nips18}, the authors propose an IP model for binary
classification, where an example is classified as positive if and only
if it satisfies at least one clause of the model.
The objective function of the IP minimizes a variation on the Hamming
loss, which is the number of incorrectly classified positive examples,
plus, for each incorrectly classified negative example, the number of
clauses incorrectly classifying it.
The complexity of the model is controlled by a bound on the size of
each clause, defined as in this paper.
Since the IP model has one binary variable for each possible clause,
the authors use column generation~\cite{barnhart-or98}.
Even so, as the pricing problem can be too expensive, it is solved
heuristically for large data sets.

\ignore{
Decision lists are sets of rules which are
hierarchically ordered by recursive if-then-else's \cite{rivest-ml87}.
Rudin and Ertekin \cite{rudin-mpc18} first build a set of rules which can overlap, and then order them.
As this is done in two separate optimization steps, there is no overall optimality guarantee, but the procedure offers other benefits such as efficiency and more flexibility in the generation of rules.
Angelino et al. \cite{angelino-kdd17} use a branch-and-bound approach with custom bounds to minimize the regularized empirical risk of the decision list.}

\section{Encoding} \label{sec:enc}

This section describes two SAT-based approaches to the problem of
computing decision sets of minimum \emph{total} size, defined as the
total number of literals used in the model.
It is useful to recall that the number of training examples is $M$,
while the number of features is $K$.
Hereinafter, it is convenient to treat a class label as an additional
feature having index $K+1$.
We first introduce models that define \emph{perfect decision sets}
that agree perfectly with the training data, and then extend these to
define \emph{sparse decision sets} that can trade off size of decision
set with classification accuracy on the training set.

\subsection{Iterative SAT Model} \label{sec:sat}

\newcommand{\noas}[1]{}
\newcommand{\newnoas}[1]{#1}

We first design a SAT model which determines whether
there exists a decision set of given size $N$.
To find the minimum $N$, we then iteratively
call this SAT model while incrementing $N$, until it is satisfied.
For every value of $N$, the problem of determining if a model of
size $N$ exists is encoded into SAT as shown below.
The idea of the encoding is that we list the rules in one after the other across
the $N$ nodes, associating a literal to each node.
The end of a rule (a \emph{leaf} node) is
denoted by a literal associated to the class.
We track which examples of the dataset are valid at each node (i.e.,
they match all the previous literals for this rule), and check that
examples that reach the end of a rule match the correct class.
The encoding uses a number of Boolean variables described below:
\begin{itemize}
   \item $s_{jr}$:$\hspace{3.3pt} $ node $j$ is a literal on feature
      $f_r\in\fml{F}\cup\fml{C}$;
   \item $t_{j}$:$\,\,$\,\,\, truth value of the literal for node $j$;
   \item $v_{ij}$:$\hspace{4.5pt} $ example $e_i\in\fml{E}$ is valid
      at node $j$;
\noas{
   \item $a_{jr}^0$:$\hspace{3.3pt} $ node $j$ assigns value 0 to feature
      $f_r\in\fml{F}\cup\fml{C}$;
   \item $a_{jr}^1$:$\hspace{3.3pt} $ node $j$ assigns value 1 to feature
      $f_r\in\fml{F}\cup\fml{C}$.
}
\end{itemize}


The model is as follows:
\begin{itemize}
   \item A node uses only one feature (or the class feature):
      \begin{equation}\label{eq:feature}
      \forall_{j \in [N]}\ \sum_{r=1}^{K+1} s_{jr} = 1
      \end{equation}
\noas{
   \item A node $j$ assigns a value 0 or 1 to feature $f_r$:
      %
      %
      \begin{equation}\label{eq:assign}
      \begin{array}{cl}
         \forall_{j\in[N]} \forall_{r\in[K+1]}\
         a_{jr}^0\leftrightarrow & (s_{jr} \land \neg{t_j}) \\
         \forall_{j\in[N]} \forall_{r\in[K+1]}\
         a_{jr}^1\leftrightarrow & (s_{jr} \land t_j) \\
      \end{array}
      \end{equation}
}
   \item The last node is a leaf:
      \begin{equation}\label{eq:last}
      s_{Nc}
      \end{equation}
   \item All examples are valid at the first node:
      \begin{equation}\label{eq:first}
      \forall_{i \in [M]}\ v_{i1} \end{equation}
   \item An example $e_i$ is valid at node $j+1$ iff $j$ is a leaf
      node, or $e_i$ is valid at node $j$ and $e_i$ and node $j$ agree
      on the value of the feature $s_{jr}$ selected for that node:
      %
      %
\noas{
      \begin{equation}\label{eq:valid}
      \forall_{i \in [M]} \forall_{j \in [N-1]}\ v_{ij+1}
      \leftrightarrow s_{jc} \vee (v_{ij} \wedge \bigvee_{r\in[K]}
      {a_{jr}^{\pi_i[r]}})
      \end{equation}
}
\newnoas{
      \begin{equation}\label{eq:valid}
      \forall_{i \in [M]} \forall_{j \in [N-1]}\ v_{ij+1}
      \leftrightarrow s_{jc} \vee (v_{ij} \wedge \bigvee_{r\in[K]}
      {(s_{jr} \wedge (t_j = {\pi_i[r]}))})
      \end{equation}
}
   \item If example $e_i$ is valid at a leaf node $j$, they should
      agree on the class feature:
      %
\noas{
      \begin{equation}\label{eq:class}
      \forall_{i \in [M]} \forall_{j \in [N]}\ (s_{jc} \wedge v_{ij})
      \rightarrow a_{jc}^{\pi_i[c]}
      \end{equation}
}
\newnoas{
      \begin{equation}\label{eq:class}
      \forall_{i \in [M]} \forall_{j \in [N]}\ (s_{jc} \wedge v_{ij})
      \rightarrow (t_j = c_i)
      \end{equation}
}
   \item For every example there should be at least one leaf literal
      where it is valid:
      \begin{equation}\label{eq:correct}
      \forall_{i \in [M]}\ \bigvee_{j\in[N]}(s_{jc} \wedge v_{ij})
      \end{equation}
\end{itemize}

The model shown above represents a non-clausal Boolean formula, which
can be clausified with the use of auxiliary
variables~\cite{tseitin68}.
Also note that any of the known cardinality encodings can be used
to represent constraint \eqref{eq:feature}~\cite[Chapter~2]{sat-handbook09}
(also see~\cite{asin-sat09,bailleux-cp03,batcher-afips68,sinz-cp05}).
Finally, the size (in terms of the number of literals) of the proposed
SAT encoding is $\fml{O}(N \times M \times K)$, which results from
constraint~\eqref{eq:valid}.

\newcommand{\true}{\mathit{true}}
\newcommand{\false}{\mathit{false}}

\begin{example}\label{ex:soln}
Consider a solution for 7 nodes for the data of \autoref{ex:data}.
The representation of the rules, as a sequence of nodes is shown below:
$$
   \newcommand{\xyo}[1]{*++[o][F-]{#1}}
    \newcommand{\xyoo}[1]{*++[o][F=]{#1}}
 \xymatrix@R=2mm{
     1  & 2 & 3 & 4 & 5& 6 & 7 \\
    \xyo{L} \ar[r] & \xyo{\neg H} & \xyo{\neg L} \ar[r] & \xyo{\neg
     C} \ar[r] & \xyo{H} & \xyo{C} \ar[r] & \xyo{\neg H}}
$$
The interesting (true) decisions for each node are given in the following
table

$$
\begin{array}{llccccccc}
   \toprule
      & \;\;\;\;\; & 1 & 2 & 3 & 4 & 5 & 6 & 7 \\ \midrule 
s_{jr} & & s_{1L} & s_{2H} & s_{3L} & s_{4C} & s_{5H} & s_{6C} & s_{7H} \\ \midrule
t_{j}  & & 1  & 0  & 0 & 0 & 1  & 1  & 0 \\ \midrule
v_{ij} & & v_{11}  & v_{12} & v_{13} & v_{34} & v_{35} & v_{16} & v_{47} \\
       & & \vdots & v_{22} & \vdots & v_{54}  &  v_{55} & \vdots  & v_{67} \\
       & &  v_{81} &  v_{42}  & v_{83} &  v_{74} &  v_{85} &  v_{86} &  v_{77} \\
       & &  &  v_{62} &  & v_{84} &  &  & \\
   \bottomrule
\end{array}
$$
Note how at the end of each rule, the selected variable is the class $H$.
Note that at the start and after each leaf node all examples are valid, and
each feature literal reduces the valid set for the next node.
In each leaf node $j$ the valid examples are of the correct class determined by
the truth value $t_j$ of that node.
\qed
\end{example}

The iterative SAT model tries to find a decision set of size $N$.
If this fails, it tries to find a decision set of size $N+1$.
This process continues until it finds a decision set of minimal size, or a time limit is reached.
The reader may wonder why we do not use binary search instead. The
difficulty with this is that the computation grows (potentially) exponentially with size
$N$, so guessing a large $N$ can mean the whole problem fails to solve.
\plb{Could it also be possible that a model of size N+1 does not exist, even if a model of size N does?}
\plb{To resolve this, would it be possible to use the same unused varibles and constraints of the maxsat version below, but solving it as a SAT? If the model is satisfied, we can inspect the solution to find a tighter N (which is not guaranteed to be minimum, but we can then run with N-1). The model with the unused variables might be easier to solve. }

\begin{example}
Consider the dataset shown in \autoref{ex:data}.
We initially try to find a decision set of size 1, which fails, then of size 2, etc.~until we reach size 7 where
we determine the decision set:
$\neg L \wedge \neg C \Rightarrow H$, $L \Rightarrow \neg H$, $C \Rightarrow
\neg H$ of size 7
by finding the model shown in \autoref{ex:soln}.
\qed
\end{example}

\subsection{MaxSAT Model} \label{sec:maxsat}

Rather than using the described iterative SAT-based procedure, which
iterates over varying size $N$ of the target decision set, we can
allocate a predefined number of nodes, which serves as an upper bound
on the optimal solution, and formulate a MaxSAT problem minimizing the
number of nodes used.
Let us add a flag variable $u_j$ for every available node.
Variable $u_j$ is \emph{true} whenever the node $j$ is unused and
\emph{false} otherwise.
%
Consider the following constraints:
\begin{enumerate}
\item A node either decides a feature or is unused:
\begin{equation}\label{eq:used} \forall_{j \in [N]}\ u_j +
\sum_{r\in[K+1]} s_{jr} = 1 \end{equation}
\ignore{ \item A node $j$ assigns a value to feature $f_r$, i.e.\
  $$
  \begin{array}{cl}
      \forall_{j\in[N]} \forall_{r\in[K+1]}\ a_{jr}^0\leftrightarrow &
      (s_{jr} \land \neg{t_j}) \\
      \forall_{j\in[N]} \forall_{r\in[K+1]}\ a_{jr}^1\leftrightarrow &
      (s_{jr} \land t_j) \\
  \end{array}
  $$ }
\item If a node $j$ is unused then so are all the following nodes
\begin{equation}\label{eq:newlast} \forall_{j \in [N-1]}\ u_{j}
\rightarrow u_{j+1} \end{equation}
\item The last used node is a leaf
\begin{eqnarray}
\forall_{j \in [N-1]}\ u_{j+1} \rightarrow u_j \vee s_{jc} \\
u_{N} \vee s_{Nc}
\end{eqnarray}
\ignore{
\item All examples are valid at the first node
$$\forall_{i \in [M]}\ v_{i1}$$
\item An example $e_i$ is valid at node $j+1$ iff $j$ is a leaf
  node, or $e_i$ is valid at node $j$ and $e_i$ and node $j$ agree
  on the value of the feature $s_{jr}$ selected for that node:
  %
  %
  $$
  \forall_{i \in [M]} \forall_{j \in [N-1]}\ v_{ij+1}
  \leftrightarrow s_{jc} \vee (v_{ij} \wedge \bigvee_{r\in[K]}
  {a_{jr}^{\pi_i[r]}})
  $$
\item If example $e_i$ is valid at a leaf node $j$, they should
  agree on the class feature, i.e.\
  %
  $$
  \forall_{i \in [M]} \forall_{j \in [N]}\ (s_{jc} \wedge v_{ij})
  \rightarrow a_{jc}^{\pi_i[c]}
  $$
\item For every example there should be at least one leaf literal
  where it is valid: $$\forall_{i \in [M]}\
  \bigvee_{j\in[N]}(s_{jc} \wedge v_{ij})$$
}
\end{enumerate}

The constraints above together with
constraints~\noas{\eqref{eq:assign}, }\eqref{eq:first},
\eqref{eq:valid}, \eqref{eq:class}, and \eqref{eq:correct} comprise
the hard part of the MaxSAT formula, i.e.\ every clause of it must be
satisfied.
As for the optimization criterion, we maximize $\sum_{j \in [N]} u_j$,
which can be trivially represented as a list of unit soft clauses of
the form $(u_j, 1)$.

The model is still used iteratively in the worst case. We guess an upper bound $N$ on the size of the decision set.
We use the model to search for a decision set of size less than or equal to $N$.  If this fails we increase $N$ by some number
(say 10) and retry, until the time limit is reached.

\begin{example}
Revisiting the solution shown in \autoref{ex:data} when $N$ is set to 9
we find the solution illustrated in \autoref{ex:soln}
extended so that the last two nodes are unused: $u_8 = u_9 = \true$.
The last used node 7 is clearly a leaf. Note that validity ($v_{ij}$) and truth
value ($t_j$) variables are irrelevant to unused nodes $j$.
\ignore{
The critical decisions are given in
the following table.
$$
\begin{array}{l|ccccccccc}
& 1 & 2 & 3 & 4 & 5 & 6 & 7 & 8 & 9 \\ \hline
s_{jr},u_j & s_{1L} & s_{2H} & s_{3L} & s_{4C} & s_{5H} & s_{6C} & s_{7H} &
u_{8} & u_9 \\
t_{j}  & \true  & \false  & \false & \false & \true   & \true  & \false  & \mbox{---} &
\mbox{---} \\
v_{ij} & v_{11}  & v_{12} & v_{13} & v_{34} & v_{35} & v_{16} & v_{47} &
v_{18} & \\
       & \vdots & v_{22} & \vdots & v_{54}  &  v_{55} & \vdots  & v_{67} &
\vdots & \\
       &  v_{81} &  v_{42}  & v_{83} &  v_{74} &  v_{85} &  v_{86} &  v_{77}
& v_{88} & \\
       &  &  v_{62} &  & v_{84} &  &  & & & \\
\end{array}
$$
Note the truth value $t_j$ and validity $v_{ij}$ is irrelevant in unused
nodes $j$, and the last used node 7 is a leaf.
}
\qed
\end{example}

\subsection{Separated Models and Multi-Classification} \label{sec:sep}

A convenient feature of minimal decision sets is the following.
The union of a minimal decision set that correctly classifies
the positive instances (and doesn't misclassify any negative instances as positive)
and a minimal decision set that correctly classifies the negative instances
(and doesn't misclassify any positive instances as negative)
is a minimal decision set for the entire problem.

We can construct a separate SAT model for the positive rules and negative rules
by simply restricting constraint~\eqref{eq:correct} to only apply to examples
in $[M]$ of the appropriate class.

Clearly, the \emph{``separated models''} are not much smaller than the complete model described in \autoref{sec:sat};
each separated model still includes constraint~\eqref{eq:valid} for each example leading to the size
$\mathcal{O}(N \times M \times K)$.
The advantage arises because the minimal size required for each half is smaller.

\begin{example}
Consider the data set shown in \autoref{ex:data}.
We can iteratively construct decision rules for the positive instances:
$\neg L \wedge \neg C \Rightarrow H$ of size 3,
and the negative instances:
$L \Rightarrow \neg H, C \Rightarrow \neg H$ of size 4.
This is faster than solving the
problems together, iterating from size 1 to size 7 to eventually find the same solution.
\qed
\end{example}

The same applies for multi-classification rules, where we need to decide on $|\cal{C}|$ different classes.
Assuming the class feature has been binarised into $|\cal{C}|$
different class binary variables,
we can modify our constraints to
build a model $M_c$ for each separate class $c \in \mathcal{C}$
as follows:

\begin{itemize}
    \item We restrict constraint~\eqref{eq:correct} to the examples $i$ in the class $c$, e.g.
      \begin{equation}\label{eq:correct-multi}
      \forall_{i \in [M], c_i = c}\ \bigvee_{j\in[N]}(s_{jc} \wedge v_{ij})
      \end{equation}
   \item We restrict leaf nodes to only consider \emph{true} examples of the class
      \begin{equation}\label{eq:true-multi}
      \forall_{j \in [N]}\ s_{jc} \rightarrow t_j
      \end{equation}
\end{itemize}
This modification is correct for both the iterative SAT and the MaxSAT models.

\subsection{MaxSAT Model for Sparse Decision Sets} \label{sec:sparse}

We can extend the MaxSAT model rather than to find minimal perfect decision sets
to look for sparse decisions sets that are accurate for most of the instances.
We minimize the objective of number of misclassifications (including non-classifications, where no decision rule gives information about the item) plus the size of the decision set
in terms of nodes multiplied by a discount factor $\Lambda$ which records that
$\Lambda$ fewer misclassifications are worth the addition of one node to the decision set.
Typically we define $\Lambda = \lceil \lambda M \rceil$ where $\lambda$ is the regularized cost
of nodes in terms of misclassifications. 

We introduce variable $m_i$ to represent that example $i \in [M]$ is misclassified.
The model is as follows:
\begin{itemize}
\item If example $e_i$ is valid at a leaf node $j$ then they agree on the
  class feature or the item is misclassified:
\noas{
      \begin{equation}\label{eq:misclassified}
      \forall_{i \in [M]} \forall_{j \in [N]}\ (s_{jc} \wedge v_{ij})
      \rightarrow a_{jc}^{\pi_i[c]} \vee m_i
      \end{equation}
}
\newnoas{
      \begin{equation}\label{eq:misclassified}
      \forall_{i \in [M]} \forall_{j \in [N]}\ (s_{jc} \wedge v_{ij})
      \rightarrow (t_j = c_i \vee m_i)
      \end{equation}
}
\item For every example there should be at least one leaf literal
  where it is valid or the item is misclassified (actually \emph{non-classified}):
  \begin{equation}\label{eq:unclassified}
  \forall_{i \in [M]}\
  m_i \vee \bigvee_{j\in[N]}(s_{jc} \wedge v_{ij})
  \end{equation}
\end{itemize}
together with all the MaxSAT constraints from \autoref{sec:maxsat} except constraints~\eqref{eq:class} and~\eqref{eq:correct}.
The objective function is
$$
\sum_{i \in [M]} m_i + \sum_{j \in [N]} \Lambda (1 - u_j) + N \Lambda
$$
represented as soft clauses $(\neg m_i, 1)$, $i\in[M]$,
and $(u_j, \Lambda)$, $j\in[N]$.

Note that the choice of regularized cost $\lambda$ is crucial.
As $\lambda$ gets higher values, the focus of the problem shifts more
to ``sparsification'' of the target decision set, instead of its
accuracy.
In other words, by selecting higher values of $\lambda$ (and hence of
$\Lambda$ as well), a user opts for simple decision sets, thus,
sacrificing their quality in terms of accuracy.
If the value of $\lambda$ is too high, the result decision set may be
empty as this will impose a high preference of the user to dispose of
all literals in the decision set.

\subsection{Separated Sparse Decision Sets} \label{sec:sparse-sep}

We can modify the definition of \emph{misclassifications} in order to support a separated solution.
Suppose that an example $e_i\in\fml{E}$ is of class $c_i\in\fml{C}$ then we count the \emph{number of misclassifications} of that example as follows:
\begin{itemize}
    \item If example $e_i$ is not classified as class $c_i$ that counts as one misclassification.
    \item If example $e_i$ is classified as class $c_j\in\fml{C}$, $c_j\neq c_i$, then this counts as one misclassification per class.
\end{itemize}
With this definition we can compute the optimal decisions sets per class independently and join them together afterwards.
The model for each class $c \in \mathcal{C}$ is identical to that of \autoref{sec:sparse} with the following change: we include constraint~\eqref{eq:true-multi} and modify constraint~\eqref{eq:unclassified} to
\begin{itemize}
\item For every example in the class $c$ there should be at least one leaf literal
  where it is valid or the example is misclassified (actually \emph{non-classified}):
  \begin{equation}\label{eq:unclassified-class}
  \forall_{i \in [M], c_i = c}\
  m_i \vee \bigvee_{j\in[N]}(s_{jc} \wedge v_{ij})
  \end{equation}
\end{itemize}
Note that there is still an $m_i$ variable for every example in every
class. For examples of class $c$ this counts as if they were \emph{not correctly classified as class $c$}, while for examples not in class $c$ it counts as if they were \emph{incorrectly classified as class $c$}.

\section{Experimental Results} \label{sec:res}
This section aims at assessing the proposed SAT-based approaches for
computing optimal decision sets from the perspective of both
scalability and test accuracy for a number of well-known datasets.

\paragraph{Experimental setup.}
The experiments were performed on the \emph{StarExec
cluster}\footnote{\url{https://www.starexec.org/}}.
Each process was run on an Intel Xeon E5-2609 2.40GHz processor with
128 GByte of memory, in CentOS~7.7.
The memory limit for each individual process was set to 16GByte.
The time limit used was set to 1800s for each individual process to
run.

\paragraph{Implementation and other competitors.}
Based on the publicly available implementation of
MinDS~\cite{ipnms-ijcar18,minds}, all the proposed models were
implemented in a prototype as a Python script instrumenting calls to
the Glucose~3 SAT solver~\cite{audemard-sat13,imms-sat18}.
The implementation targets the models proposed in the paper, namely,
(1)~the iterative SAT model studied in \autoref{sec:sat} and (2)~its
MaxSAT variant (see \autoref{sec:maxsat}) targeting minimal perfect
decision set but also (3)~the MaxSAT model for computing sparse
decision sets as described in ~\autoref{sec:sparse}.
All models target independent computation of each class\footnote{The
prototype adapts all the developed models to the case of multiple
classes, which is motivated by the practical importance of non-binary
classification.}, as discussed in \autoref{sec:sep} and
\autoref{sec:sparse-sep}.
As a result, the iterative SAT model and its MaxSAT variant in the
following are referred to as \emph{opt} and \emph{mopt}.
Also, to illustrate the advantage of separated models over the model
aggregating all classes, an aggregated SAT model was tested, which is
referred to as \emph{opt$_\cup$}.
Finally, several variants of the MaxSAT model targeting sparse
decision sets called \emph{sp$[\lambda_i]$} were tested with three
values of regularized cost: $\lambda_1=0.005$, $\lambda_2=0.05$, and
$\lambda_3=0.5$.
One of the considered competitors were MinDS$_2$ and
MinDS$_2^\star$~\cite{ipnms-ijcar18}, which in the following are
referred to as \emph{mds$_2$} and \emph{mds$_2^\star$}, respectively.
While the former tool minimizes the number of rules, the latter does
lexicographic optimization, i.e.\ it minimizes the number of rules
first and then the total number of literals.
Additionally, MinDS was modified to produce \emph{sparse} decision
sets, similarly to what is described in Section~\ref{sec:sparse}.
This extension also makes use of MaxSAT to optimize the sparse
objective rather than SAT, which was originally used.
In the following comparison, the corresponding implementation is named
\emph{mds$_2[\rho_i]$}, with regularization cost $\rho_1=0.05$,
$\rho_2=0.1$, and $\rho_3=0.5$.
Note that $\rho_i\neq\lambda_i$, $i\in[3]$ since the measures used by
the two models are different.
One targets rules and the other -- literals.
In order to, more or less, fairly compare the scalability of the new
model \emph{sp} and of the sparse variant of \emph{mds$_2^\star$}, we
considered a few configurations of \emph{sp$[\rho_i]$} with
$\rho_i=\frac{\lambda_i}{K}$, where $K$ is the number of features in a
dataset, where we consider a rule equivalent to $K$ literals.
To tackle the MaxSAT models, the RC2-B MaxSAT solver was
used~\cite{imms-jsat19}.

\begin{figure*}[!t]
  \begin{subfigure}[b]{0.49\textwidth}
    \centering
    \includegraphics[width=\textwidth]{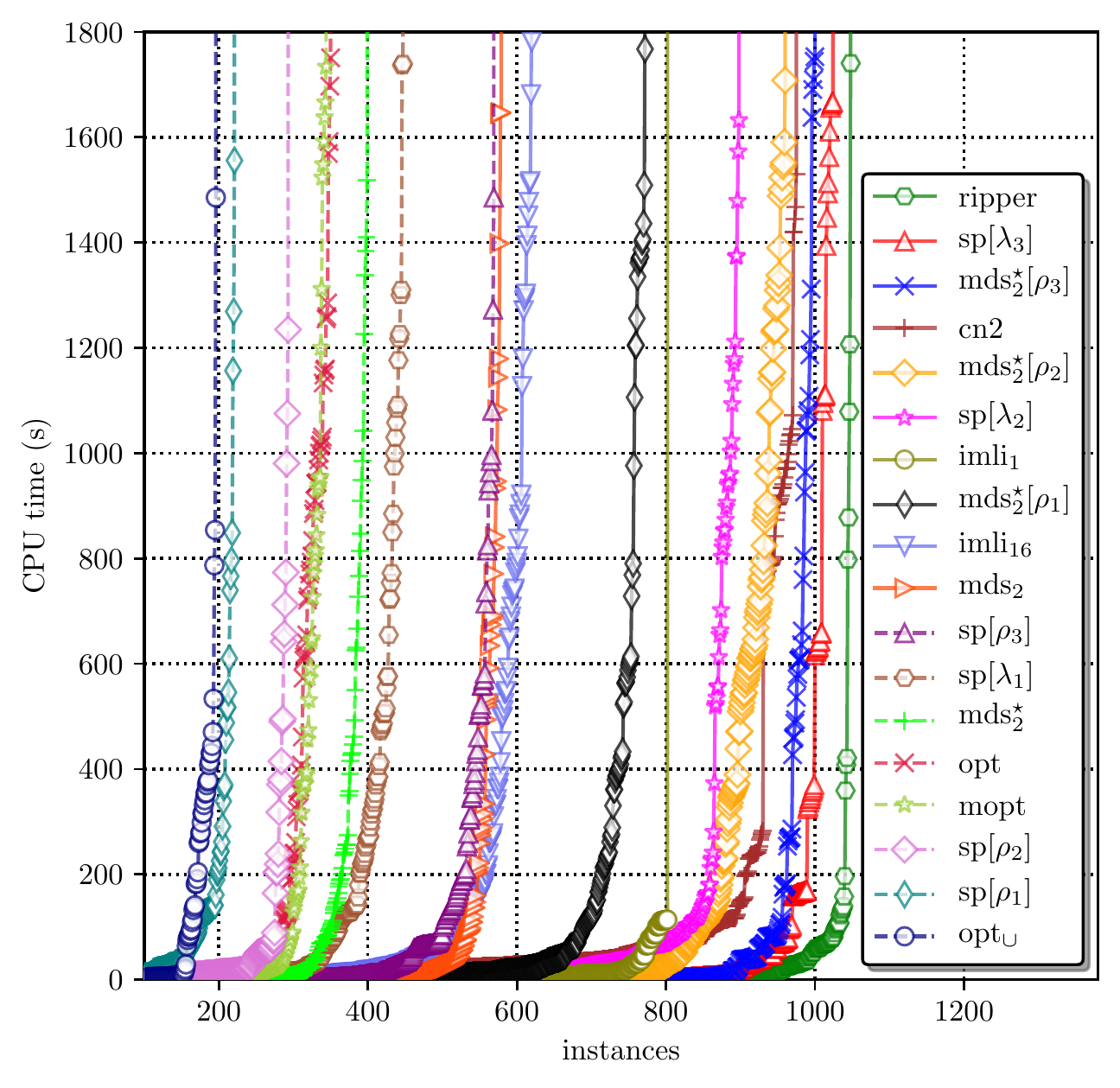}
    \caption{Raw performance}
    \label{fig:perf}
  \end{subfigure}%
  \begin{subfigure}[b]{0.49\textwidth}
    \centering
    \includegraphics[width=\textwidth]{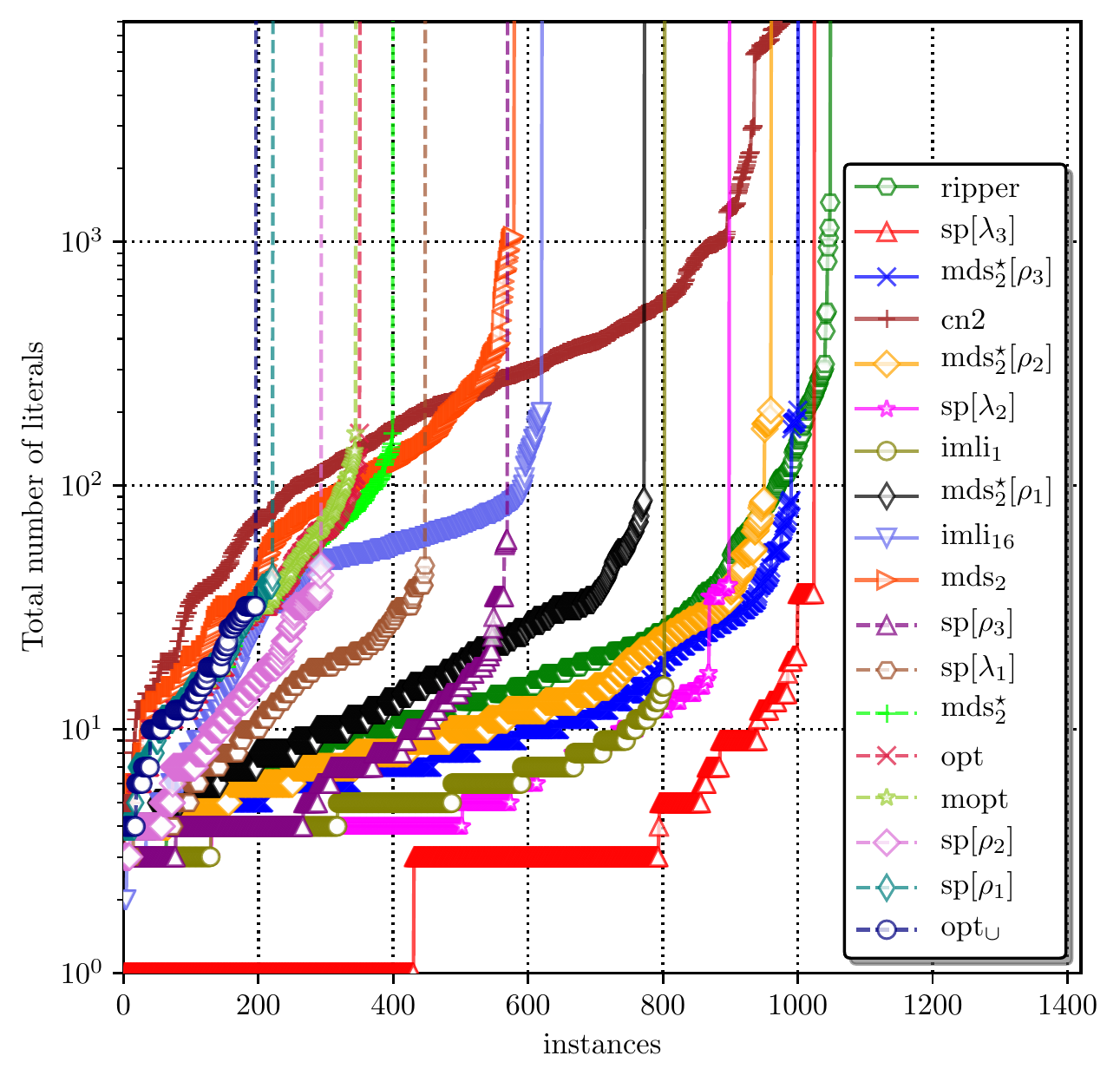}
    \caption{Total number of literals used}
    \label{fig:size}
  \end{subfigure}
  \caption{Scalability of all competitors on the complete set of instances
    and the quality of solutions i terms of decision set size.}
  \label{fig:full}
\end{figure*}

A number of state-of-the-art algorithms were additionally considered
including the heuristic methods CN2~\cite{clark-ml89,clark-ewsl91},
and RIPPER~\cite{cohen-icml95}, as well as MaxSAT-based
IMLI~\cite{meel-aies19} (which is a direct successor of
MLIC~\cite{meel-cp18}).
The implementation of CN2 was taken from Orange~\cite{orange} while a
publicly available implementation of RIPPER~\cite{wittgenstein} was
used.
It should be noted that given a training dataset, IMLI and RIPPER
compute only one class.
To improve the accuracy reported by both of these competitors, we used
a \emph{default rule} that selects a class (1)~different from the
computed one and (2)~represented by the majority of data instances in
the training data.
The default rule is applied only if none of the computed rules can be
applied.
Finally, IMLI takes a constant value $k$ of rules in the clausal
representation of target class to compute.
We varied $k$ from $1$ to $16$.
The best results (both in terms of performance and test accuracy) were
shown by the configuration targeting the smallest possible number of
rules, i.e.\ $k=1$; the worst results were demonstrated for $k=16$.
Thus, only these extreme values of $k$ were used below represented by
\emph{imli$_1$} and \emph{imli$_{16}$}, respectively.

\begin{figure*}[!t]
  \begin{subfigure}[b]{0.49\textwidth}
    \centering
    \includegraphics[width=\textwidth]{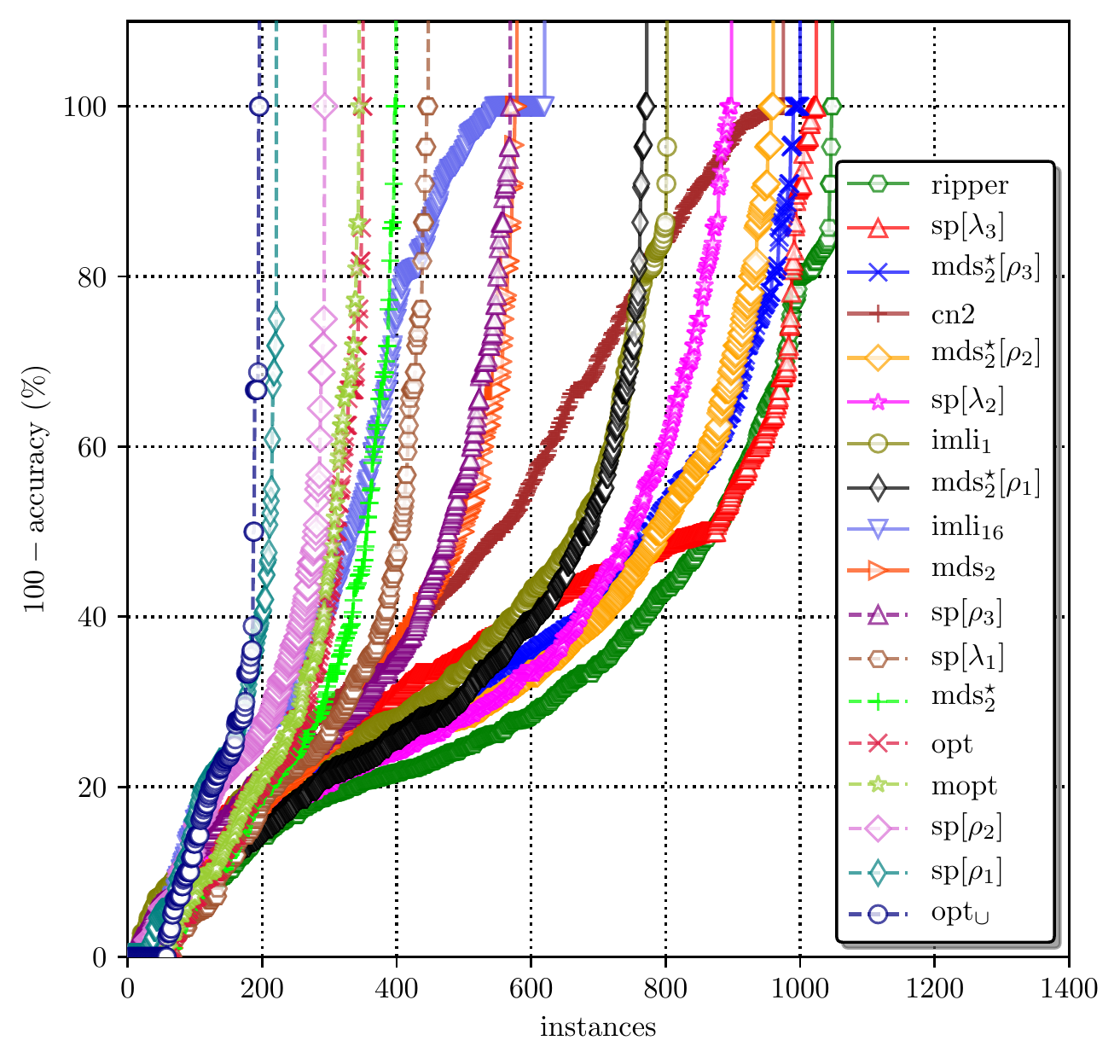}
    \caption{On the complete set of datasets}
    \label{fig:accy-full}
  \end{subfigure}%
  \begin{subfigure}[b]{0.49\textwidth}
    \centering
    \includegraphics[width=\textwidth]{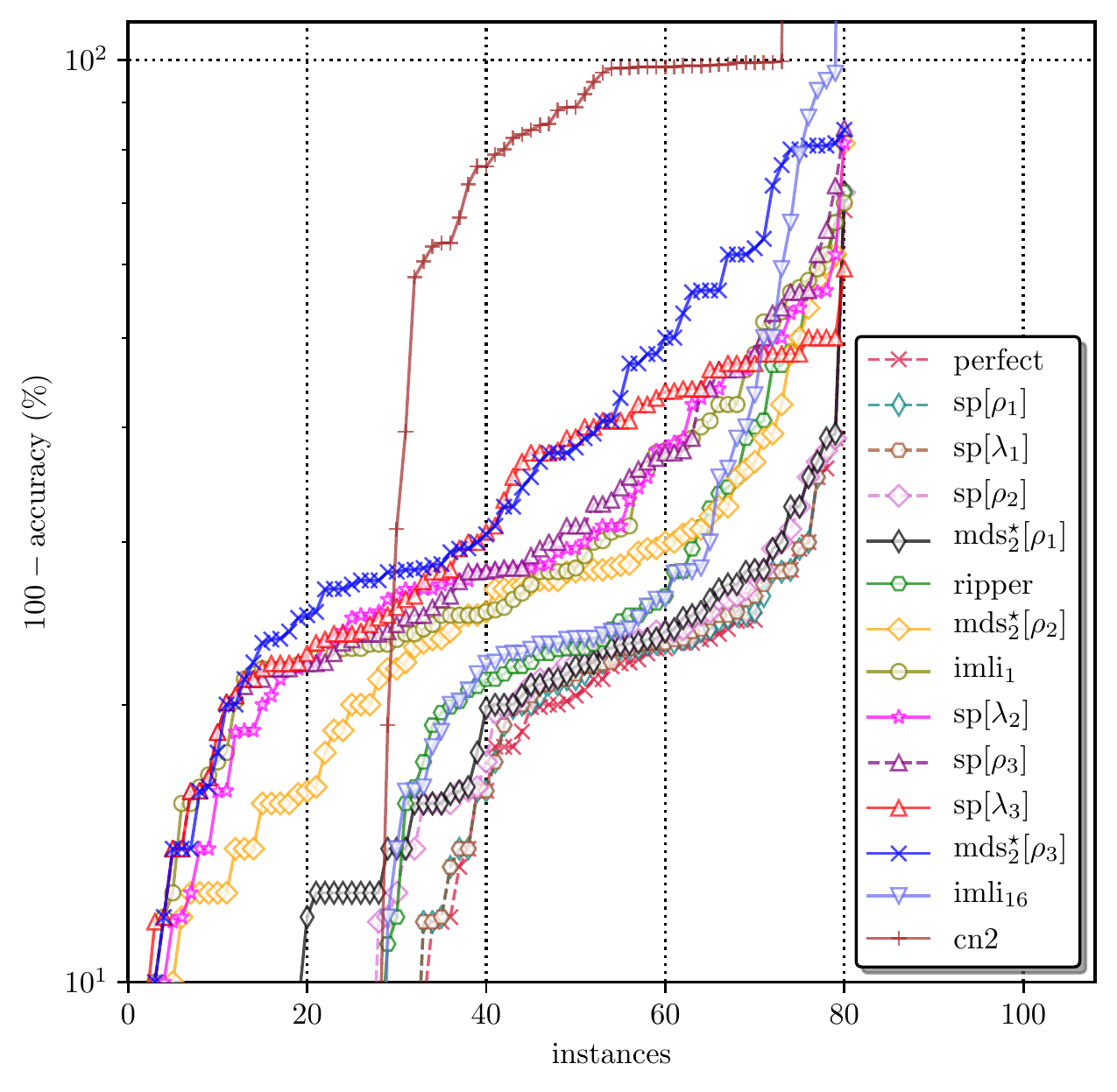}
    \caption{Instances solved by all tools}
    \label{fig:accy-simp}
  \end{subfigure}
  \caption{Accuracy of the considered approaches.}
  \label{fig:accy}
\end{figure*}

\paragraph{Datasets and methodology.}
Experimental evaluation was performed on a subset of datasets selected
from publicly available sources.
These include datasets from UCI Machine Learning Repository~\cite{uci}
and Penn Machine Learning Benchmarks.
Note that all the considered datasets were previously studied
in~\cite{ipnms-ijcar18,meel-aies19}.
The number of selected datasets is 71.
Ordinal features in all datasets were quantized so that the domain of
each feature gets to $2$, $3$, or $4$.
This resulted in 3 families of benchmarks, each of size 71.
Whenever necessary, quantization was followed by \emph{one-hot
encoding}~\cite{scikitlearn}.
In the datasets used, the number of data instances varied from 14 to
67557 while the number of features after quantization varied from 3 to
384.

Finally, we applied the approach of 5-fold cross validation,
i.e.\ each dataset was randomly split into 5 chunks of instances; each
of these chunks served as test data while the remaining 4 chunks were
used to train the classifiers.
This way, every dataset (out of 71) resulted in 5 individual pairs of
training and test datasets represented by 80\% and 20\% of data
instances.
Therefore, each quantized family of datasets led to 355 pairs of
training and test datasets.
Hence, the total number of benchmark datasets considered is 1065.
Every competitor in the experiment was run to compute a decision set
for each of the 1065 training datasets, which was then tested for
accuracy on the corresponding test data.

It is important to mention that the accuracy for all the tools was
tested by an external script in a unified way.
Concretely, (1)~if a rule \emph{``covers''} a data instance of a wrong
class, the instance is deemed misclassified (even if there is another
rule of the right class covering this data instance);
(2)~if none of the rules of a given class covers a data instance of
that class, the instance is deemed misclassified.
Afterwards, assuming the total number of misclassified instances is
denoted by $E$ while the total number of instances is $M$, the
accuracy is computed as a value $\frac{M-E}{M} \times 100\%$.

\begin{figure*}[!t]
  \begin{subfigure}[b]{0.49\textwidth}
    \centering
    \includegraphics[width=\textwidth]{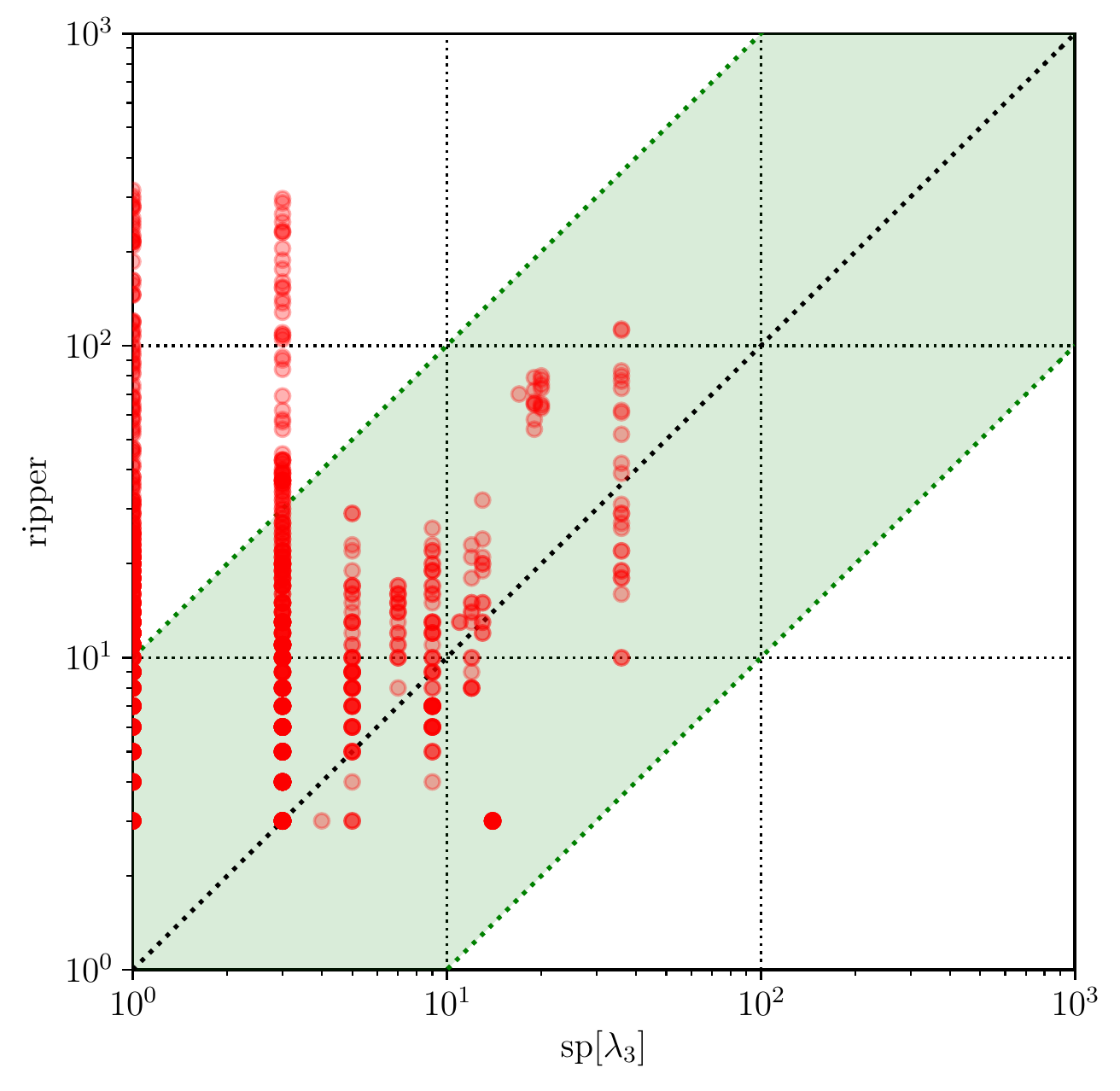}
    \caption{Solution quality}
    \label{fig:rip-size}
  \end{subfigure}%
  \begin{subfigure}[b]{0.49\textwidth}
    \centering
    \includegraphics[width=\textwidth]{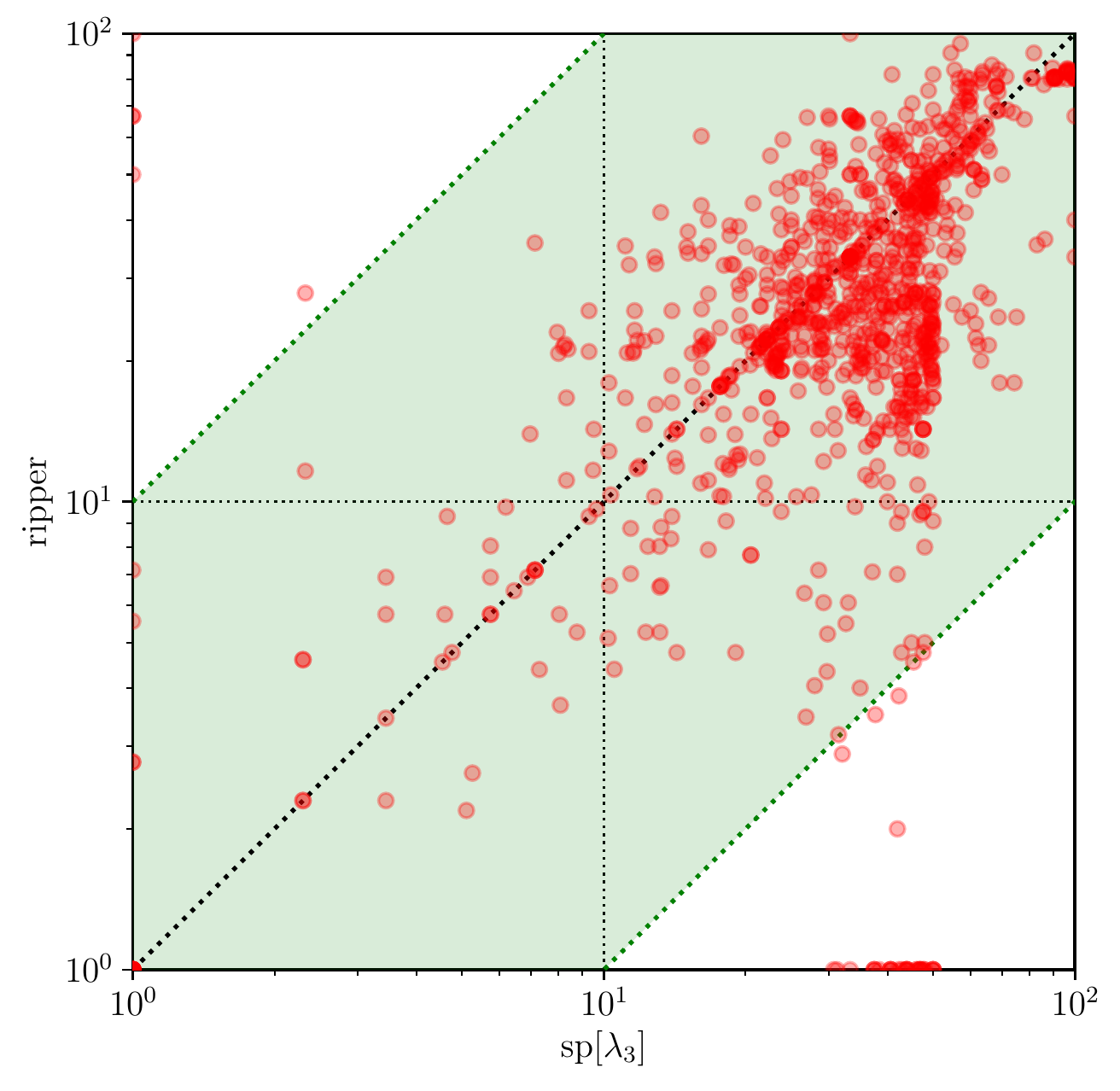}
    \caption{Accuracy}
    \label{fig:rip-accy}
  \end{subfigure}
  \caption{Comparison of sp$[\lambda_3]$ and ripper.}
  \label{fig:vsripper}
\end{figure*}

\paragraph{Testing scalability.}
\autoref{fig:perf} shows the performance of the considered competitors
on the complete set of benchmark instances.
As one can observe, \emph{ripper} outperforms all the other tools and
is able to train a classifier for 1048 of the considered datasets
given the 1800s time limit.
The proposed MaxSAT models for sparse decision sets
\emph{sp}$[\lambda_3]$ and \emph{mds$_2^\star[\rho_3]$} (which are the
configurations with the largest constant parameters) come second and
third with 1024 and 1000 instances solved, respectively.
The fourth place is taken by \emph{cn2}, which can successfully deal
with 975 datasets.
The best configuration of \emph{imli}, i.e.\ \emph{imli$_1$}, finishes
with 802 intances solved while the worst configuration
\emph{imli$_{16}$} copes with only 620 datasets.
Finally, the worst results are demonstrated by the approaches that
target \emph{perfectly accurate} decision sets \emph{opt},
\emph{mopt}, and \emph{opt$_\cup$} but also by the sparse approaches
with low regularized cost \emph{sp$[\rho_1]$} and
\emph{sp$[\rho_2]$}.
For instance, \emph{opt$_\cup$} solves only 196 instances.
This should not come as surprise since the problem these tools target
is computationally harder than what the other approaches solve.

\paragraph{Testing accuracy.}
Having said that, perfectly accurate decision sets once computed have
the highest possible accuracy.
This is confirmed by \autoref{fig:accy-simp}, which depicts the
accuracy obtained by all the tools for the datasets solved by
\emph{all} the tools.
Indeed, as one can observe, the virtual \emph{perfect} tool, which
acts for all the approaches targeting perfectly accurate decision
sets, i.e.\ \emph{opt}, \emph{mopt}, \emph{opt$_\cup$},
\emph{mds$_2$}, and \emph{mds$_2^\star$}, beats the other tools in
terms of test accuracy.
Their average test accuracy on these datasets is
$85.89\%$\footnote{This average value is the highest possible
accuracy that can be achieved on these datasets whatever machine
learning model is considered.}.
In contrast, the worst accuracy is demonstrated by \emph{cn2}
($43.73\%$).
Also, the average accuracy of \emph{ripper}, \emph{sp$[\lambda_3]$},
\emph{mds$_2^\star[\rho_3]$}, \emph{imli$_1$}, \emph{imli$_{16}$} is
$80.50\%$, $67.42\%$, $61.71\%$, $76.06\%$, $77.42\%$, respectively.

The picture changes drastically if we compare test accuracy on the
complete set of benchmark datasets.
This information is shown in \autoref{fig:accy-full}.
Here, if a tool does not solve an instance, its accuracy for the
dataset is assumed to be $0\%$.
Observe that the best accuracy is achieved by \emph{ripper} ($68.13\%$
on average) followed by the sparse decision sets computed by
\emph{sp$[\lambda_3]$} and \emph{mds$_2^\star[\rho_3]$} ($60.91\%$ and
$61.23\%$, respectively).
The average accuracy achieved by \emph{imli$_1$} and
\emph{imli$_{16}$} is $50.66\%$ and $28.26\%$ while the average
accuracy of \emph{cn2} is $47.49\%$.

\begin{figure*}[!t]
  \begin{subfigure}[b]{0.49\textwidth}
    \centering
    \includegraphics[width=\textwidth]{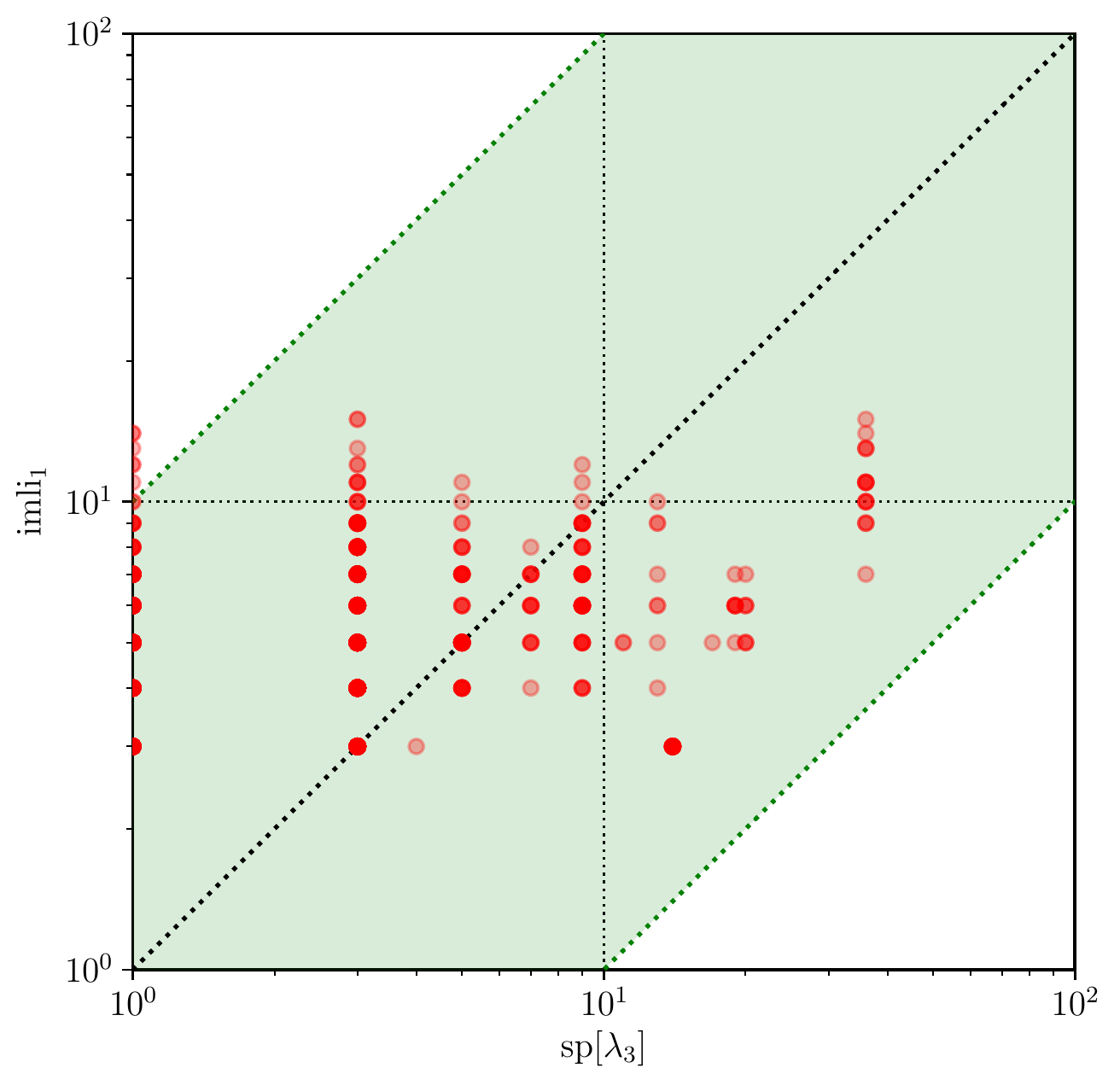}
    \caption{Solution quality}
    \label{fig:imli-size}
  \end{subfigure}%
  \begin{subfigure}[b]{0.49\textwidth}
    \centering
    \includegraphics[width=\textwidth]{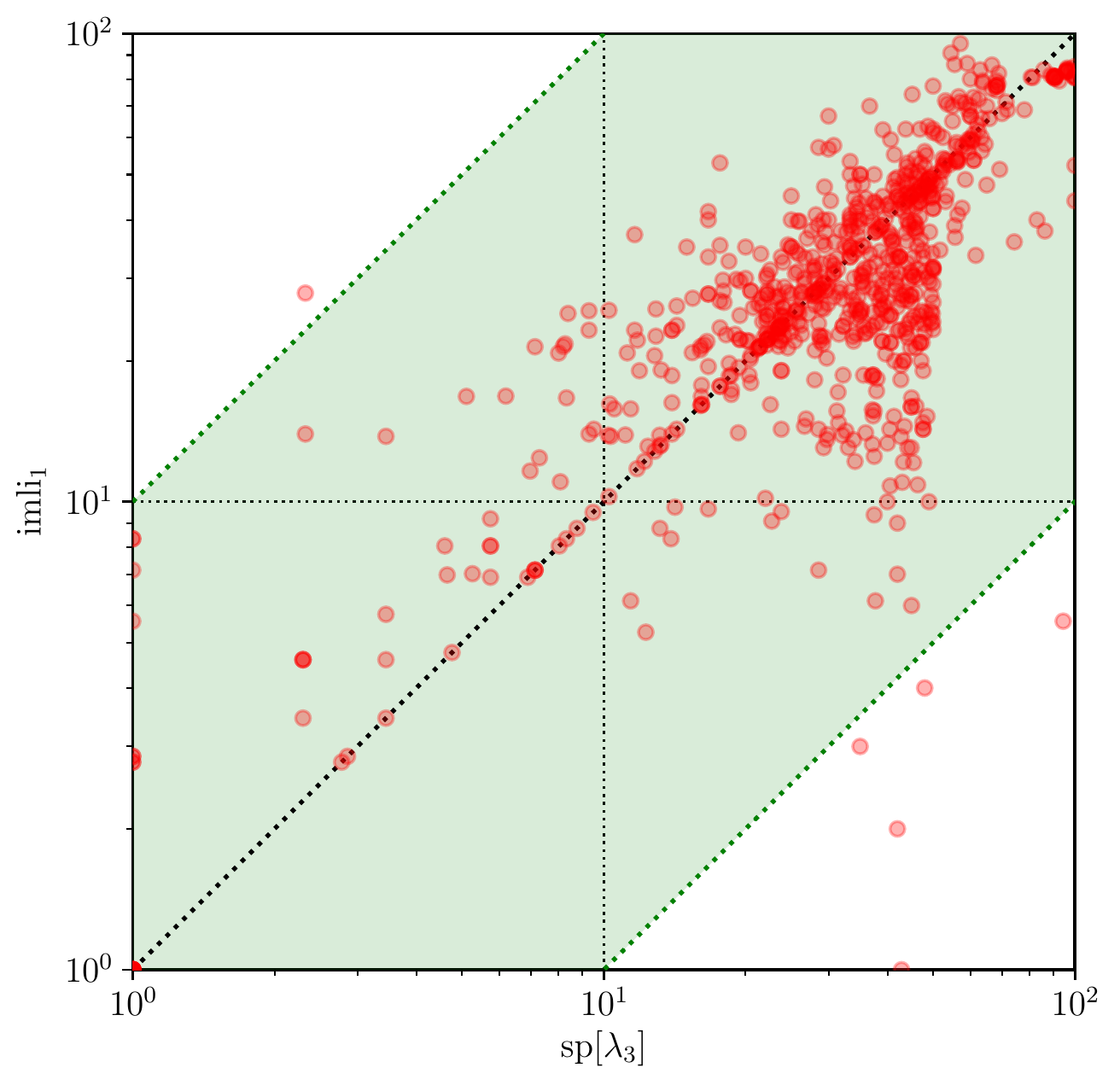}
    \caption{Accuracy}
    \label{fig:imli-accy}
  \end{subfigure}
  \caption{Comparison of sp$[\lambda_3]$ and imli$_1$.}
  \label{fig:vsimli}
\end{figure*}

\paragraph{Testing interpretability (size).}
From the perspective of interpretability, the smaller a decision set
is the easier it is for a human decision maker to comprehend.
This holds for the number of rules in a decision set but also (and
more importantly) for the total number of literals used.
\autoref{fig:size} depicts a cactus plots illustrating the size of
solutions in terms of the number of literals obtained by each of the
considered competitors.
A clear winner here is \emph{sp$[\lambda_3]$}.
As can be observed, for more than 400 datasets, decision sets of
\emph{sp$[\lambda_3]$} consist of only one literal\footnote{In a
unit-size decision set, the literal is meant to assign a constant
class. This can be seen as applying a \emph{default rule}.}.
Another bunch of almost 400 datasets are represented by
\emph{sp$[\lambda_3]$} with 3 literals.
Getting these small decision sets is a striking achievement in light
of the overall high accuracy reached by \emph{sp$[\lambda_3]$}.
The average size of solutions obtained by \emph{sp$[\lambda_3]$} is
$4.18$.
Note that \emph{imli$_1$} gets close to this with $5.57$ literals per
dataset on average although it always compute only one rule.
In clear contrast with this, the average solution size of
\emph{ripper} is $35.14$ while the average solution of
\emph{imli$_{16}$} has $46.29$ literals.
Finally, the result of \emph{cn2} is $598.05$ literals.

It is not surprising that perfectly accurate decision sets, i.e.\
those computed by \emph{opt}, \emph{opt$_\cup$}, \emph{mopt}, as well
as \emph{mds$_2$} and \emph{mds$_2^\star$}, in general tend to be
larger.
It is also worth mentioning that \emph{mds$_2^\star[\rho_3]$} obtains
sparse decision sets of size $14.52$ on average while the original
(non-sparse) version gets $40.70$ literals per solution.

\paragraph{A few more details.}
\autoref{fig:vsripper} and \autoref{fig:vsimli} detail a comparison of
\emph{sp$[\lambda_3]$} with \emph{ripper} and \emph{imli$_1$},
respectively.
All these plots are obtained for the datasets solvable by each pair of
competitors.
Concretely, as can be seen in \autoref{fig:imli-size} and
\autoref{fig:imli-accy}, the size and accuracy of
\emph{sp$[\lambda_3]$} and \emph{imli$_1$} are comparable.
However, as \emph{imli$_1$} computes solutions representing only one
class and it is significantly outperformed by \emph{sp$[\lambda_3]$},
the latter approach is deemed a better alternative.
Furthermore and although the best performance overall is demonstrated
by \emph{ripper}, its accuracy is comparable with the accuracy of
\emph{sp$[\lambda_3]$} (see \autoref{fig:rip-accy}) but the size of
solutions produced by \emph{ripper} can be several orders of magnitude
larger than the size of solutions of its rival, as one can observe in
\autoref{fig:rip-size}.

Finally, a crucial observation to make is that since both RIPPER and
IMLI compute a representation for one class only, they cannot provide
a user with a succinct explanation for the instances of \emph{other
(non-computed) classes}.
Indeed, an explanation in that case includes the negation of the
complete decision set.
This is in clear contrast with our work, which provides a user with a
succinct representation of every class of the dataset.

\section{Conclusion} \label{sec:conc}%

We have introduced the first approach to build decision sets by
directly minimizing the total number of literals required to describe
them.
The approach can build perfect decision sets that match the training
data exactly, or sparse decision sets that trade off accuracy on
training data for size.
Experiments show that sparse decision sets can be preferred to
perfectly accurate decision sets.
This is caused by (1)~their high accuracy overall and (2)~the fact
that they are much easier to compute.
Second, it is not surprising that the regularization cost
significantly affects the efficiency of sparse decision sets -- the
smaller the cost is, the harder it is compute the decision set and the
more accurate the result decision set is.
This fact represents a reasonable trade-off that can be considered in
practice.
Note that points 1 and 2 hold for the models proposed in this paper
but also for the \emph{sparse variants} of prior work targeting
minimization of the number of rules~\cite{ipnms-ijcar18}.
Third, although heuristic methods like RIPPER may scale really well
and produce accurate decision sets, their solutions tend to be much
larger than sparse decision sets, which makes them harder to
interpret.
All in all, the proposed approach to sparse decision sets embodies a
viable alternative to the state of the art represented by prior
logic-based solutions~\cite{ipnms-ijcar18,meel-cp18,meel-aies19} as
well as by efficient heuristic
methods~\cite{cohen-icml95,clark-ml89,clark-ewsl91}.

There are number of interesting directions to extend this work.
There is considerable symmetry in the models we propose, and while we tried
adding symmetry breaking constraints to improve the models, what we tried
did not make a significant difference. This deserves further exploration.
Another interesting direction for future work is to consider other measures of
interpretability, for example the (possibly weighted) average length of a decision rule,
where we are not concerned about the total size of the decision set, but
rather its succinctness in describing any particular instance.

\subsection*{Acknowledgments}

This work is partially supported by the Australian Research Council through
Discovery Grant DP170103174.

\bibliographystyle{abbrv}
\bibliography{refs}

\end{document}